\documentclass[USenglish,twocolumn]{article}
\usepackage[utf8]{inputenc}
\usepackage[big,online]{dgruyter}
\usepackage[sort&compress,square,numbers]{natbib}
\usepackage{times}
\usepackage{siunitx}
\usepackage{microtype}
\usepackage{todonotes}
\graphicspath{{./figures/}{./logos}}
\usepackage{subfig}
\usepackage[outline]{contour}
\usetikzlibrary{arrows.meta}

\usepackage{blindtext}

\usepackage[detect-all=true,separate-uncertainty=true, multi-part-units=single,range-phrase={-},]{siunitx}

\tikzset{%
  dblarw/.style n args={2}{%,
    -{Circle[scale=0.6]},
    line width=1.5pt,
    draw=#2,  % this draw and color could
    color=#2, % be an additional style arg
    opacity=1,
    postaction={
      draw=#1,
      color=#1,
      line width=1pt,
      shorten >=0.5pt,
      shorten <=0.3pt,
    },
    dblarw/.default={black}{white},
    dblarw/.initial={black}{white},
  }
}

\tikzset{%
  coordarw/.style n args={2}{%,
    -stealth,
    line width=1.5pt,
    draw=#2,  % this draw and color could
    color=#2, % be an additional style arg
    opacity=1,
    postaction={
      draw=#1,
      color=#1,
      line width=1pt,
      shorten >=0.5pt,
      shorten <=0.3pt,
    },
    dblarw/.default={black}{white},
    dblarw/.initial={black}{white},
  }
}

\begin{document}

%%%--------------------------------------------%%%
%%% Please do not alter the following lines: %%%
%%%--------------------------------------------%%%
%\articletype{Proceedings}
%\aop
\DOI{10.1515/}
\openaccess
\pagenumbering{gobble}
%%%--------------------------------------------%%%

\title{VR-based body tracking to stimulate musculoskeletal training  }
% VR-based training for whole body activation
% VR for unsupervised whole body training
% Can head movement alone predict whole body movement for VR based training?
% Towards adaptive VR based training for whole body movement
% An adaptive VR game for guiding and evaluating physical training
\runningtitle{VR based Body Tracking}
\author*[1]{M. Neidhardt}
\author*[1]{S. Gerlach}
\author[2]{F. N. Schmidt}
\author[2]{I. A. K. Fiedler}
\author[3]{S. Grube}
\author[2]{B. Busse}
\author[3]{A. Schlaefer}
\runningauthor{M. Neidhardt et al.}

\affil[1]{\protect\raggedright 
Hamburg University of Technology, Institute of Medical Technology and Intelligent Systems, Hamburg, Germany \newline email: maximilian.neidhardt@tuhh.de \newline *Both authors contributed equally.}
\affil[2]{\protect\raggedright 
University Medical Center Hamburg-Eppendorf, Department of Osteology and Biomechanics, Hamburg, Germany}
\affil[3]{\protect\raggedright 
Institute of Medical Technology and Intelligent Systems, Hamburg University of Technology, Hamburg, Germany}

\abstract{
Training helps to maintain and improve sufficient muscle function, body control, and body coordination. These are important to reduce the risk of fracture incidents caused by falls, especially for the elderly or people recovering from injury. Virtual reality training can offer a cost-effective and individualized training experience. We present an application for the HoloLens 2 to enable musculoskeletal training for elderly and impaired persons to allow for autonomous training and automatic progress evaluation. We designed a virtual downhill skiing scenario that is controlled by body movement to stimulate balance and body control. By adapting the parameters of the ski slope, we can tailor the intensity of the training to individual users. In this work, we evaluate whether the movement data of the HoloLens 2 alone is sufficient to control and predict body movement and joint angles during musculoskeletal training. We record the movements of 10 healthy volunteers with external tracking cameras and track a set of body and joint angles of the participant during training. We estimate correlation coefficients and systematically analyze whether whole body movement can be derived from the movement data of the HoloLens 2. No participant reports movement sickness effects and all were able to quickly interact and control their movement during skiing. Our results show a high correlation between HoloLens 2 movement data and the external tracking of the upper body movement and joint angles of the lower limbs. %% more

}


\maketitle
\section{Introduction} 

\begin{figure*}[htb]
    \centering
    \subfloat[]{
    \begin{tikzpicture} 
        \node[inner sep=0,outer sep=0] (image) at (0,0) {\includegraphics[height=3.7cm]{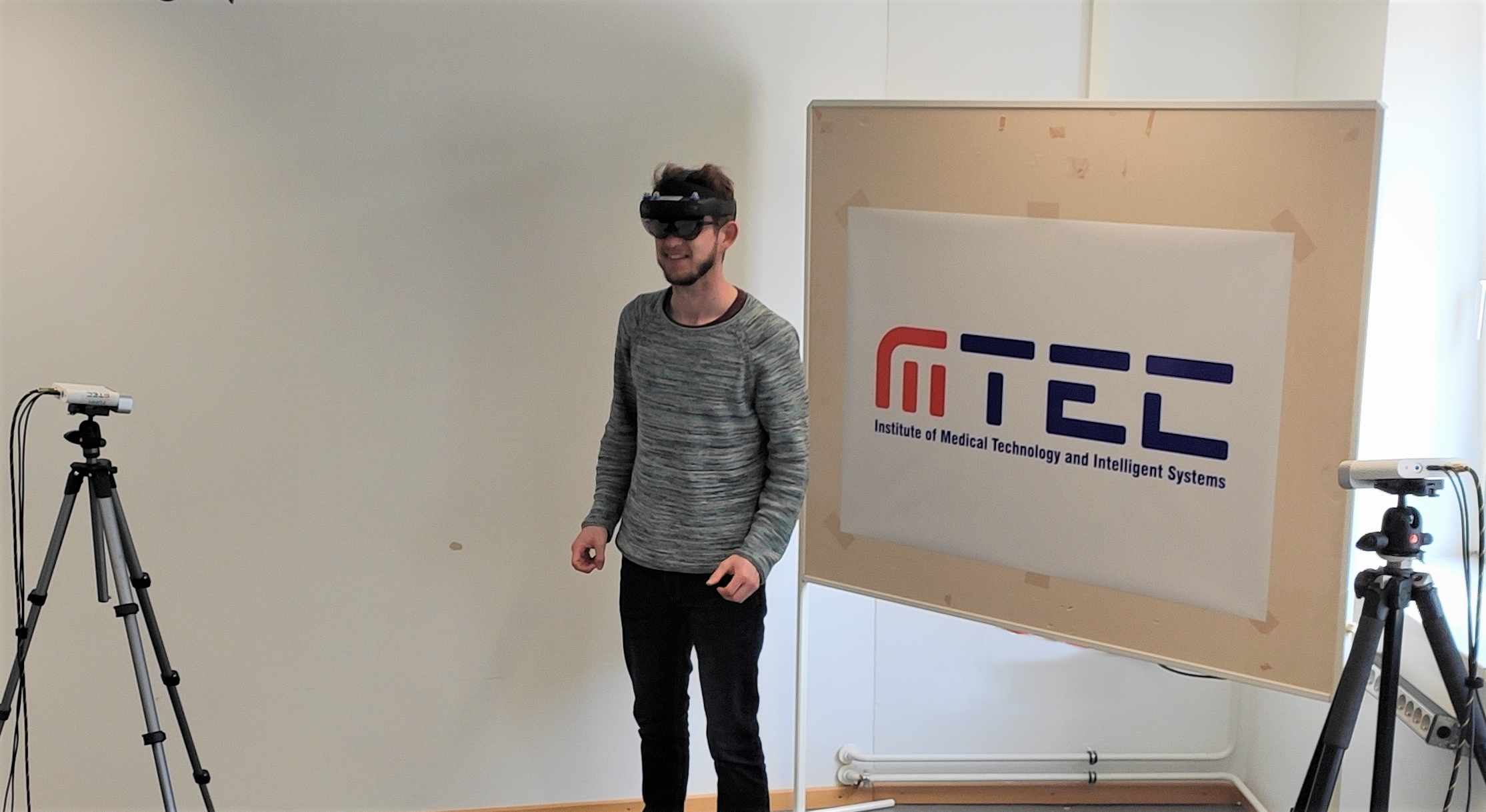}};
        \draw[dblarw={red}{white}] (-2,-1.5) node[rectangle,draw=red,line width=0.1mm,inner sep=0.5mm,preaction={fill, white},above] {\textcolor{red}{Tracking Camera 1}} -- (-3.08,0.08);
        \draw[dblarw={red}{white}] (2,1) node[rectangle,draw=red,line width=0.1mm,inner sep=0.5mm,preaction={fill, white},above] {\textcolor{red}{Tracking Camera 2}} -- (3.15,-0.32);
        \draw[dblarw={red}{white}] (-1.5,1.4) node[rectangle,draw=red,line width=0.1mm,inner sep=0.5mm,preaction={fill, white},above] {\textcolor{red}{HoloLens 2}} -- (-0.4,0.9);
        \draw[coordarw={red}{white}] (-0.3,0.82) -- (-1.1,0.56) node[rectangle,draw=red,line width=0.1mm,inner sep=0.5mm,preaction={fill, white},below left] {\textcolor{red}{z}};
        \draw[coordarw={red}{white}] (-0.3,0.82) -- (.4,0.68) node[rectangle,draw=red,line width=0.1mm,inner sep=0.5mm,preaction={fill, white},below right,yshift=-2.] {\textcolor{red}{x}};
        \draw[coordarw={red}{white}] (-0.3,0.82) -- (-0.3,1.62) node[rectangle,draw=red,line width=0.1mm,inner sep=0.5mm,preaction={fill, white},right,xshift=4] {\textcolor{red}{y}};
    \end{tikzpicture}
        \label{fig:expSetupA}
    }
    \hfill
    \subfloat[]{
        \includegraphics[height=3.7cm]{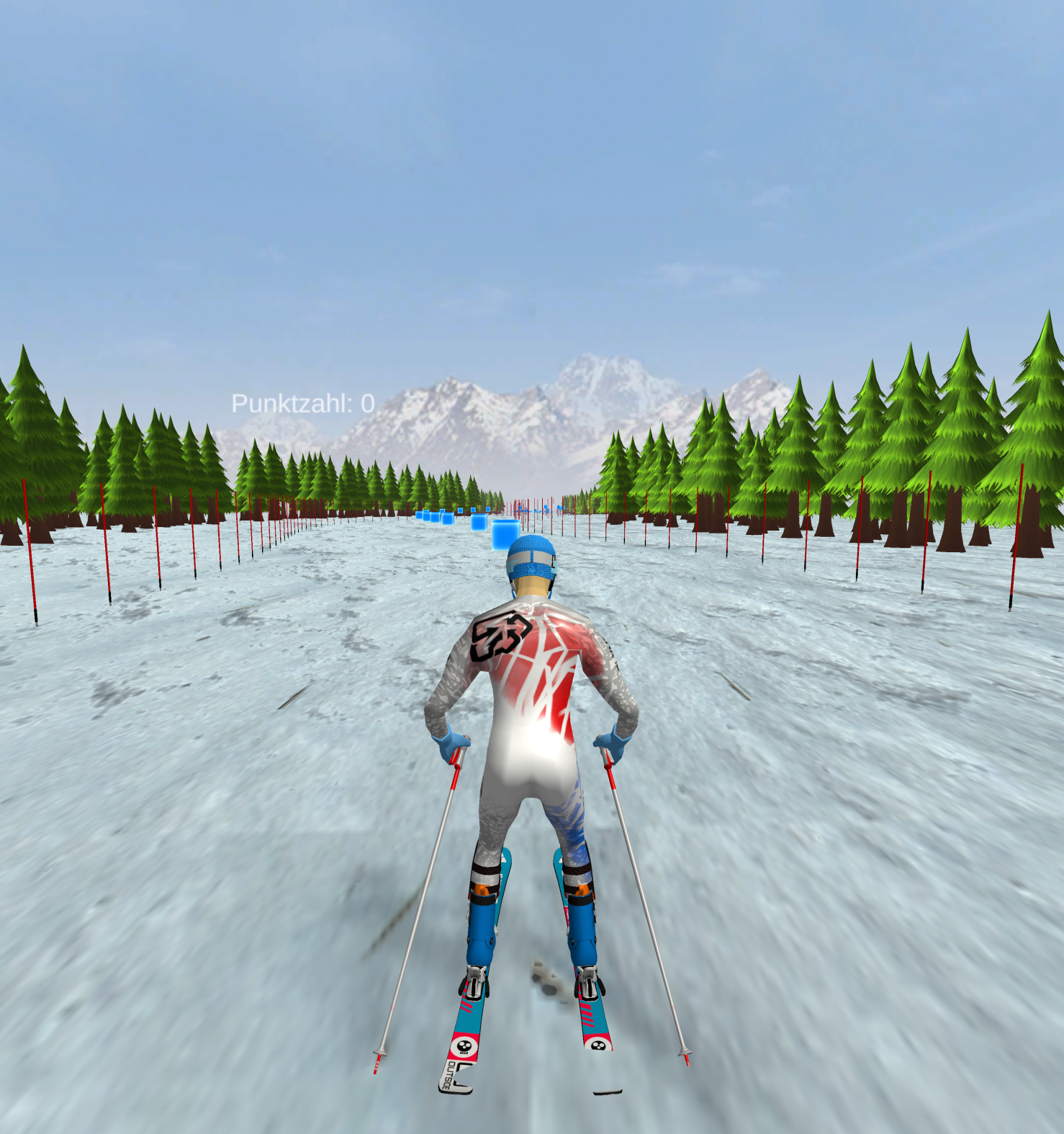}
        \includegraphics[height=3.7cm]{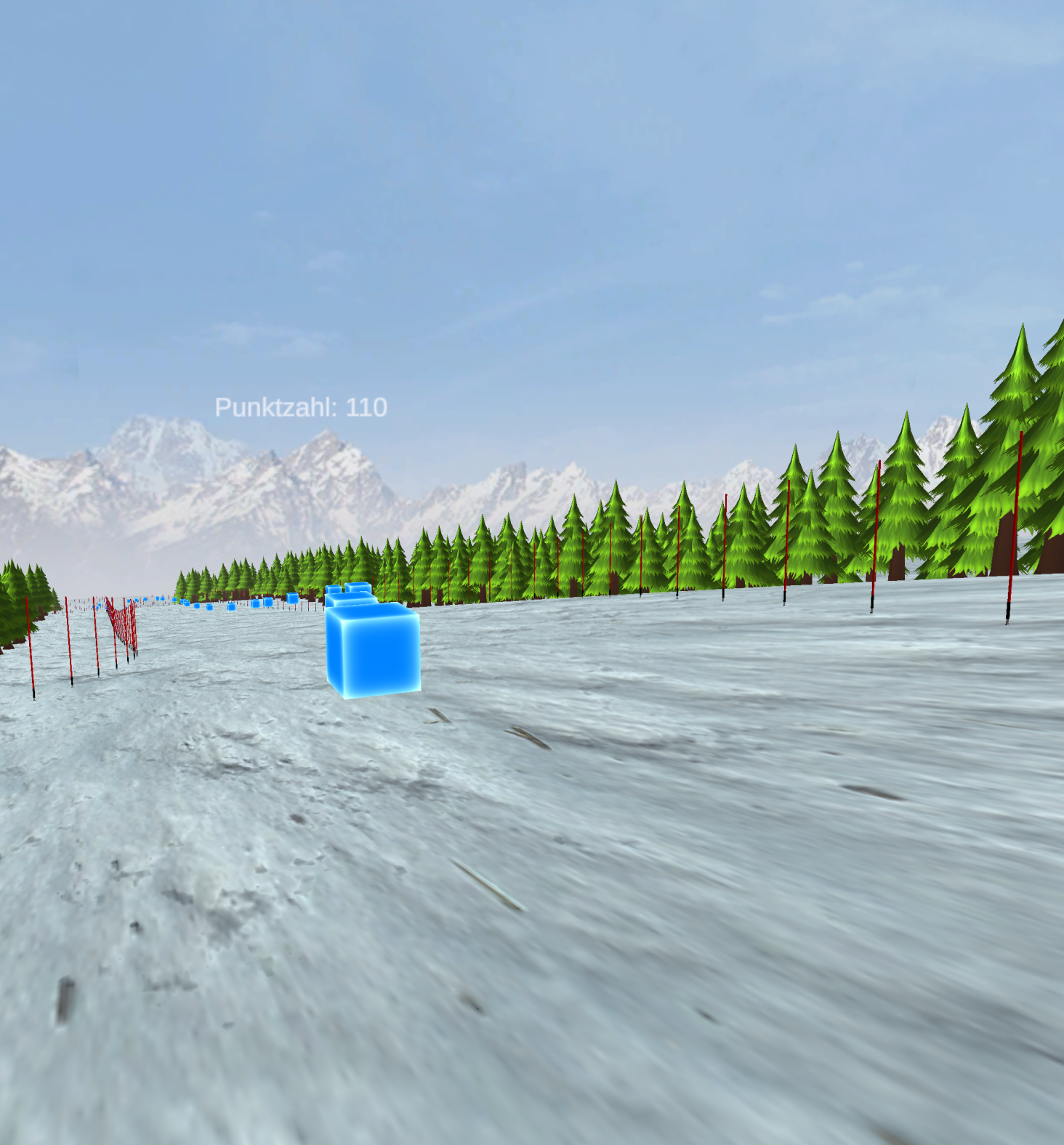}
        \label{fig:expSetupB}
    }
        \hfill
    \subfloat[]{
        \includegraphics[height=3.7cm]{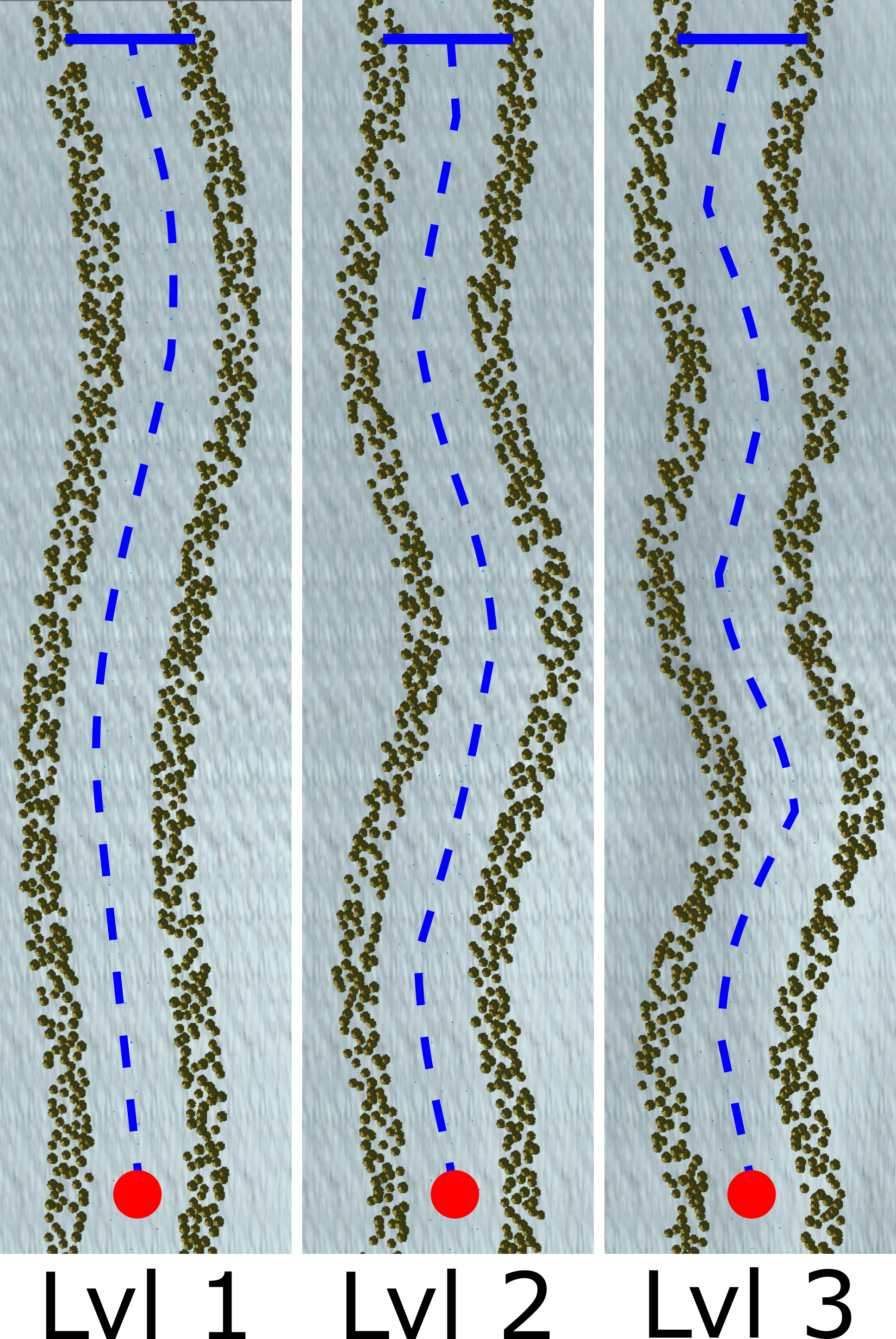}
        \label{fig:expSetupC}
    }
    \caption{\textbf{Experimental setup for data acquisition:} (a) user with Hololens during training while two cameras track body movement. Tracking cameras are only used for the experimental evaluation and are not needed for the application itself. (b) In-game view for the user with (left) and without (right) a virtual player model. The poles and trees delimit the slope. The user can earn points by skiing through blue boxes. (c) Top view of levels with low (Lvl 1), medium (Lvl 2) and high (Lvl 3) difficulty. Indicated are start (red dot), goal (blue line) and trajectory (dashed blue line). As the level increases turns get tighter and the terrain becomes more challenging.}
    \label{fig:expSetup}
\end{figure*}
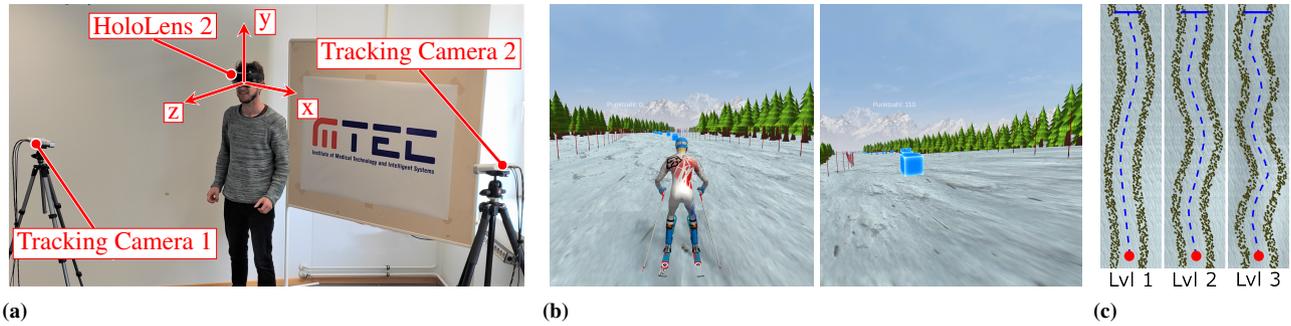

%\begin{figure*}[htb]
%\subfloat[Level 1 (Easy)]{\includegraphics[width=0.32\linewidth]{Level1}}
%\hfill
%\subfloat[Level 2 (Medium)]{\includegraphics[width=0.32\linewidth]{Level2}}
%\hfill
%\subfloat[Level 3 (Challenging)]{\includegraphics[width=0.32\linewidth]{Level3}}
%\caption{Top view of levels with varying difficulty. The trees and poles limit the track. Additionally to tighter turns, the terrain gets more challenging with increasing level.}
%\label{fig:levels}
%\end{figure*}

Life expectancy has risen sharply in recent decades, but the demographic change poses a challenge for medical care \cite{Whitty.2020}. Especially in rural areas, telemedicine offers an opportunity for efficient care where the physician-to-patient ratio is unfavorable, the distances to the nearest physician are long and patients might be impaired. Moderate regular physical exercise provides a range of health benefits, especially for old people \cite{Vogel.2009} or people recovering from an injury, e.g. brain injury \cite{Devine.2009}.

Telemedicine via virtual reality (VR) or augmented reality (AR) has become more accessible in recent years due to the commercial availability of VR and AR glasses. We propose to use the HoloLens 2 (Microsoft, USA) for a cost-effective, controlled form of physical training for elderly and impaired patients. VR provides a valuable benefit in combination with conventional physical training \cite{Mirelman.2011, Schoene.2013, Vovk.2018}. Especially balance exercises which do not include walking have been shown to prevent falls in older people \cite{Sherrington.2008}. However, an adaption of the training to individual patients is crucial since old and recovering patients can exhibit a wide range of capabilities for body movement and therefore require personalized training \textcolor{black}{which has rarely been done with only head-mounted devices \cite{Zahabi.2020}}. %\textcolor{red}{To this end we investigate if we can derive body motion purely from tracking data recorded by the internal sensors of the HoloLens 2. Thereby, training scenarios can be adapted to patients with a low-cost and simple setup.}

In this work, we implement a virtual downhill skiing scenario as a physical exercise on the HoloLens 2. The movement of the virtual player character is controlled only by head movement which is recorded by the HoloLens 2. Thereby, our setup is simple to use for home training and no external devices are needed. We calibrate the control of the virtual player character according to the capability for body movement of the user. In this work, we investigate if we can predict distinct body movements such as joint angles from the movement data recorded by the HoloLens 2. Therefore, we track the body movement during the experiments with a separate tracking camera setup. We estimate the correlation of distinct body movements from 10 healthy subjects with internal tracking data recorded by the HoloLens 2. We also show how we can control the type and range of movement of the user by parameterizing the downhill skiing scenario. Thereby we demonstrate that individualized training is feasible with minimal active human supervision.

\vspace{-0.4cm}
\section{Methods} 

\begin{figure}[htb]
    \centering
    \includegraphics[height=3.4cm]{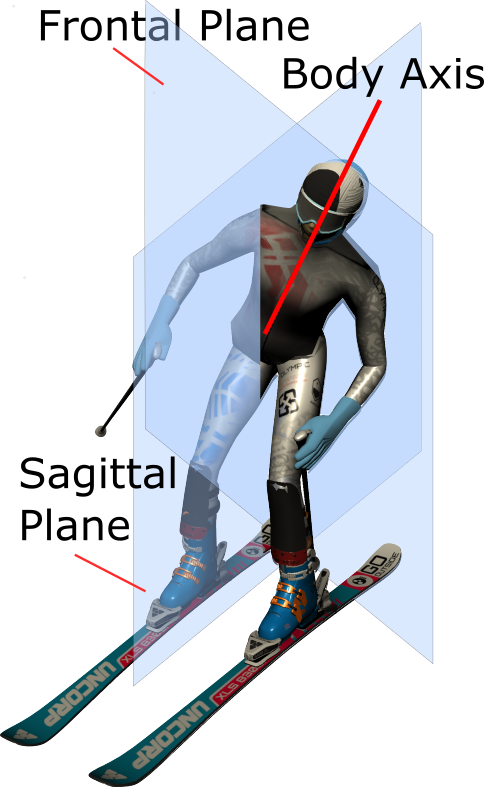}
    \hspace{1.2cm}
    \includegraphics[height=3.4cm]{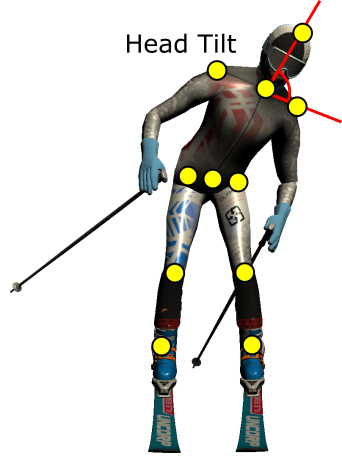}\\
    \includegraphics[height=3.4cm]{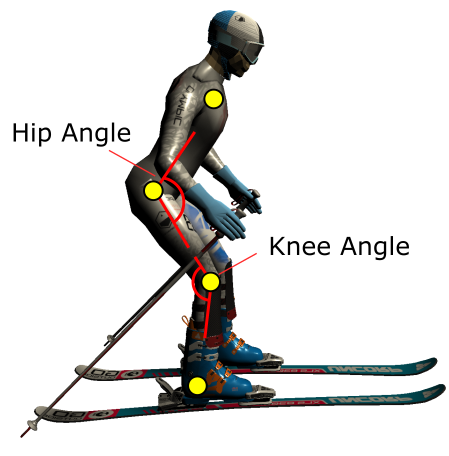}
    \hspace{1.2cm}
    \includegraphics[height=3.4cm]{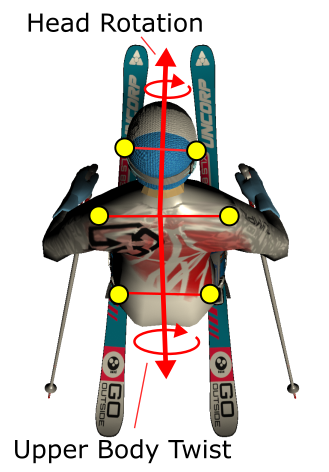}
    \caption{\textbf{Body model tracking:} Indicated are in red the angles and the yellow dots depict the position of the fitted body model.}
    \label{fig:jointAngles}
\end{figure}

\begin{figure*}[htb]
    \centering
    \subfloat[]{
        \includegraphics[height=5.7cm]{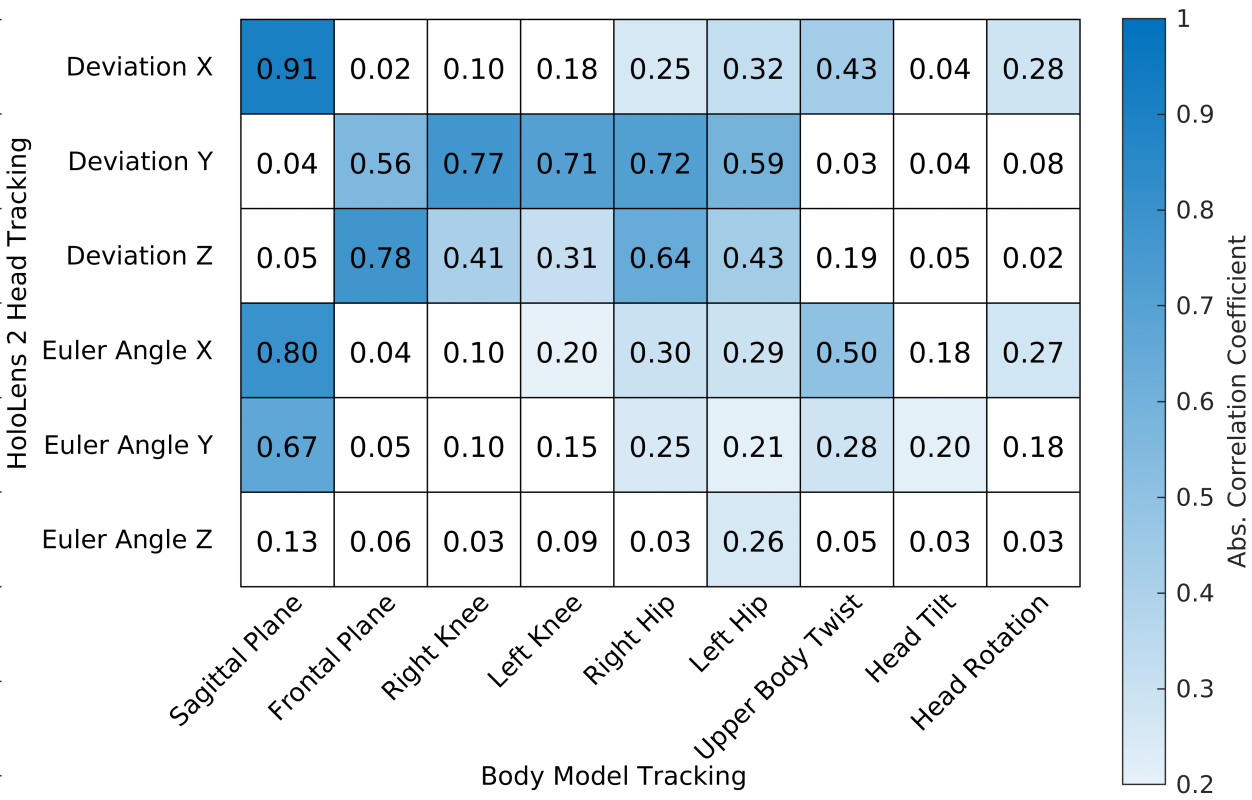}
        \label{fig:ResultsA}
    }
    \hfill
     \subfloat[]{
        \includegraphics[height=5.7cm]{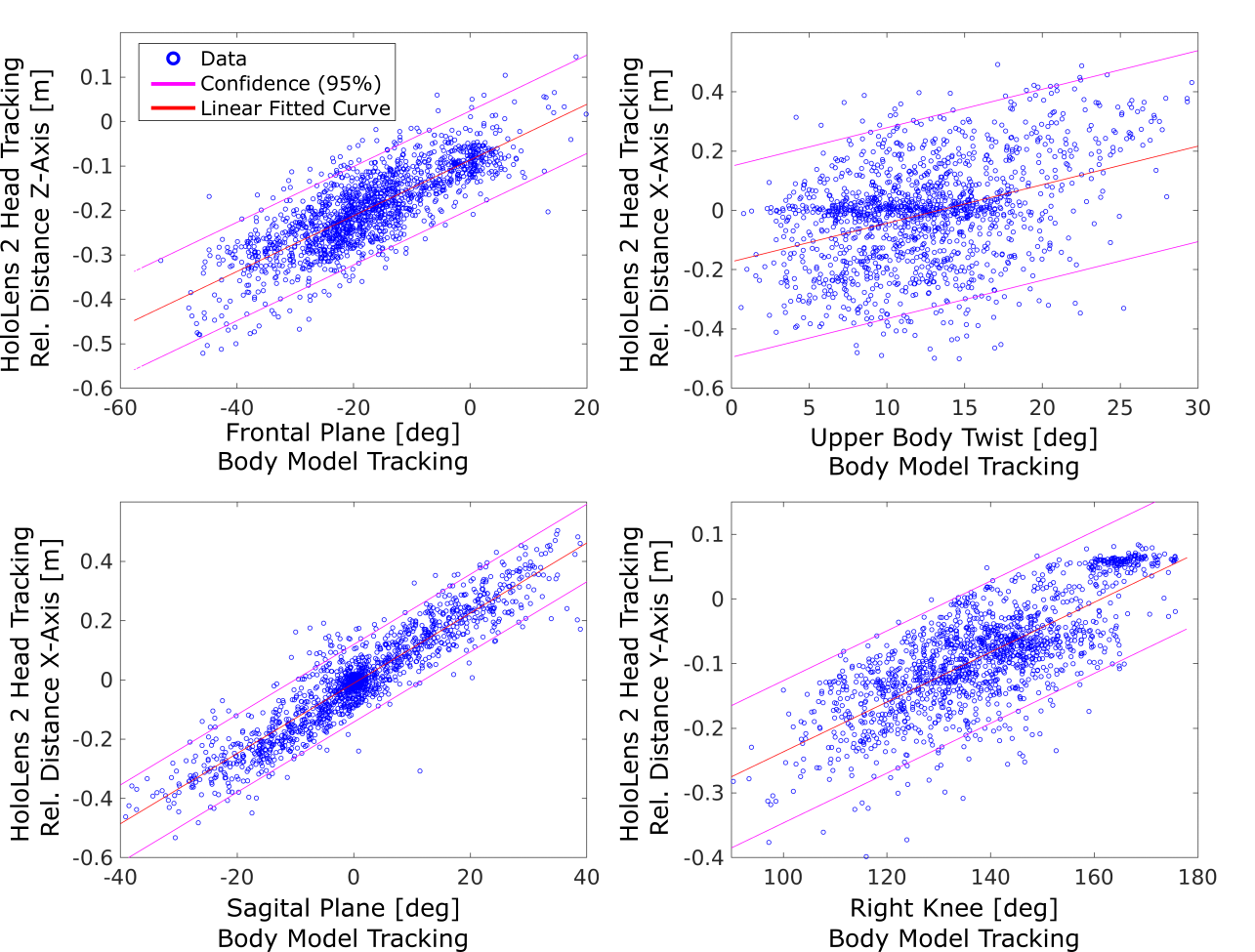}
        \label{fig:ResultsB}
    }   
    \hfill
    \caption{\textbf{Experimental Results:} (a) Absolute Correlation coefficient between the tracked position of the HoloLens 2 and the recorded body model angles (all experiments: P<0.01). (b) Scatter plots with indicated 95\% prediction interval (pink) for selected body model angles and recorded HoloLens 2 head positions. }
    \label{fig:Results}
\end{figure*}

\subsection{VR Application}
We implement a downhill skiing scenario in the Unity game engine (Unity Technologies, USA). The HoloLens 2 can track 6 degrees of freedom of the head movement by its integrated depth camera and inertial measurement unit. 
During the startup of the application, we first calibrate the range of movement of the user. We prompt the user to take a skiing stance and lean to the sides and back and front. Here, we guide the user by showing a virtual player model that performs the desired movements. We track the head movement and calculate the movement range.

We generate the terrain height map by starting with a linearly descending terrain. We add 2D sampled random noise and interpolate intermediate height values. We limit the slope which is reachable by the player with poles and trees and add blue cubes in the middle of the slope as shown in Figure~\ref{fig:expSetupB}. These cubes reward the player with points and motivate the player to execute the desired body movement.

When skiing, the virtual player model is simulated as a physical object using the physics engine of Unity. A head movement along the X-axis of the HoloLens 2 shown in Figure~\ref{fig:expSetupA} results in a momentum applied to the player model to rotate left/right, respectively. Thereby, the player can initiate curves to follow the track. Player head movement \textcolor{black}{along the Z-axis adapts the speed at which the player moves.} To ensure that the player is in a skiing position and to increase the training effect, we control that the head position is below the upright position of the player by an offset as calibrated in the beginning. When the player stands upright, the speed and rotational momentum are reduced.

\subsection{Experimental Setup}
Additionally to the HoloLens 2 which is running the skiing application, we use two tracking cameras (Microsoft Azure Kinect, Microsoft, USA) to record the patient movements as depicted in Figure~\ref{fig:expSetupA}. We track the body movement in real-time with the Microsoft body tracking SDK\footnote{https://www.microsoft.com/en-us/download/details.aspx?id=44561} which is widely used for markerless body tracking. Based on the depth-resolved and RGB images, a skeleton with 32 joints is aligned to the body movement with a framerate of approximately \SI{25}{\Hz}. We select one tracking camera for each joint movement, depending on the visibility of the joint.

\subsection{Data Acquisition and Processing}
Each participant is given a short introduction to the handling of the HoloLens 2. The participants are given the task of keeping their feet in a fixed position during the experiment while controlling movements in the game purely through upper body movement. First, \textcolor{black}{a calibration procedure is performed where the user is tasked to lean to the left, right, front, and back such that they can still stand comfortably. The maximum deviations in the positions are recorded and used to linearly scale the players motion during training.} For data acquisition, the participant performs three different levels depicted in Figure~\ref{fig:expSetupC}. The difficulty changes depending on the level by decreasing the radii of the sampled curves, increasing the steepness of the terrain and noise amplitude. Moreover, we evaluate if a virtual player model which depicts the movement has an effect on the body movement of the participant.

We track the body movement with two external tracking cameras as shown in Figure~\ref{fig:expSetupA}. We define nine distinct body joint angles which are depicted in Figure~\ref{fig:jointAngles}. Before each level the user stands in an upright position. In this position, we estimate the frontal and sagittal plane and subsequently record the enclosed angle of the body axis and the planes during the training. We correlate the change of the body angles during an exercise run with the recorded head position by the HoloLens 2. Joint positions estimated with low confidence by the tracking camera are excluded.

\subsection{Participants}
We evaluate our system on 10 healthy volunteers. The mean age of the group is \SI{28.8(29)} with a 70:30 male-to-female ratio. On average, each participant performed \SI{3(1)} days of physical activity during the week and considered themselves to be active in sports with \SI{5.5(0.7)} points on a scale from 0 to 10. 70\% of the participants had very limited experience with a HoloLens 2 and 30\% had fair experience before data acquisition. All participants gave written informed consent prior to data acquisition.

\vspace{-0.4cm}
\section{Results and Discussion}
None of the participants reported motion sickness during data acquisition. Similar to results reported in the literature, the usage of the HoloLens 2 did not cause motion sickness \cite{Vovk.2018}. In general, the use of the headset during training was well tolerated (5-7 minutes), and users quickly grasped the interaction with the software and movement control. However, our study group is young, healthy and physically active. Whether motion sickness or limited interaction with the application occurs with elderly or impaired users needs to be further evaluated.

The correlation between the tracked position of the head reported by the HoloLens 2 and the external tracking camera is shown in Figure~\ref{fig:ResultsA}. We report a very high correlation \cite{Mukaka.2012} between the deviation along the X-axis and the sagittal plane angle and a high correlation for the angle enclosed with the frontal plane and the deviation along the Z-axis. A moderate correlation is given for the movement along the Y-axis with the knee and hip angle. Based on the head tracking with the HoloLens 2 the overall upper body movement can be derived from the movement along X- and Z-axis, as well as the Euler angles of the X- and Y-axis. Exemplary absolute estimates of the angle and the reported position of the HoloLens 2 are depicted in Figure~\ref{fig:ResultsB}. Considering these results, we can predict the movement of the upper body and to some extent lower body movement by considering head movement alone. More complex models could further improve these predictions.

% In general I would assume here that the Euler Angle around Y-Axis would somehow correlate with the head rotation. Maybe it is strange to externals that the Head rotation and tilt can not be estimated from the HoloLens 2 (Head set). 
Only a low correlation is found for the head movement. Interestingly, the head rotation, which we define relative to the shoulder axis, did not correlate with the Euler angle around Y-axis. This might be due to the upper body twist during training and the limited tracking accuracy of the shoulder position.

We evaluate the effect of level creation parameters on the training intensity. The maximum tracked body angles increase with more difficult levels as shown in Table~\ref{tab:MinMaxJoints}. This indicates that more extreme movements are necessary to complete levels of higher difficulty. Additionally, the total distance which the head travels during training increases with increasing difficulty of levels as shown in Table~\ref{tab:HeadVsAvatar}. This is due to the increased number of curves in the trajectory for higher levels and a generally higher number of corrections required by the user to stay on track. Our study results did not show increased movements of the head when a virtual player model depicts the user's input movement. However, the virtual player model might reduce the learning curve for elderly and not experienced users. \textcolor{black}{In addition, the acceptance of the HoloLens 2 as a training device needs to be further studied for impaired or elderly patients.}

%%%%%%%%%%%%%%%%%%%%%%%%%%%%%%%%%%%%%%%%%%%%%%%%%%%%%%%%%%%%
%% Points Reached  %%
%   user | lvl 1 No Avatar | lvl 1 Avatar | lvl 2 No Avatar | lvl 2 Avatar | lvl 3 No Avatar | lvl 3 Avatar
%   1 300   330   170   200   200   140
%   2 270   390   110   280   150    30
%   3 360   390   370   370   320    80
%   4 260   360   270   250   190   300
%   5 390   360   210   300   220   260
%   6 390   390   300    90   170   170
%   7 390   390   350   280   310   170
%   8  40   180    70    30    60   240
%   9 380   290   390   350   380   270
%   10 380   300   370   350   310   120
%%%%%%%%%%%%%%%%%%%%%%%%%%%%%%%%%%%%%%%%%%%%%%%%%%%%%%%%%%%%

\begin{table}
\setlength\tabcolsep{-0.0cm}
\caption{Average and standard deviation of maximum movement of joint angles. Deg is degree; pp is percentage points.}
\begin{tabular*}{0.484\textwidth}{
@{\extracolsep{\fill}}
  l
  S[table-format=-2.2(4)]
  S[table-format=-2.2(4)]
  S[table-format=-2.2(4)]
  @{}
}
Body Model & {$\text{Level 1}$} & {$\text{Level 2}$} & {$\text{Level 3}$}\\ \midrule
Sagittal Plane [deg] & 43.46\pm14.60 & 49.46\pm14.34  & 56.33\pm6.67\\
Frontal Plane [deg] & 39.56\pm13.11 & 35.21\pm14.47  & 37.21\pm10.55 \\
Right Knee [deg] & 46.85\pm13.83 & 52.03\pm14.67  & 56.41\pm11.99 \\
Left Knee [deg] & 49.36\pm14.89 & 52.62\pm16.46  & 53.50\pm15.59\\
Right Hip [deg] & 45.34\pm12.35 & 47.97\pm11.67  & 51.98\pm11.64\\
Left Hip [deg] & 35.49\pm8.64 & 38.48\pm10.42  & 41.03\pm10.07\\
Up. Body Twist [deg] & 20.09\pm6.09 & 22.90\pm7.84  & 25.29\pm5.98\\
Head Tilt [deg] & 54.64\pm18.39 & 51.80\pm18.33  & 57.38\pm15.13\\
Head Rotation [deg] & 51.71\pm13.13 & 51.29\pm16.26  & 54.12\pm15.63\\ \midrule
Deviation to Lv1 [pp]    &  \text{Ref.}  &   4.74\pm8.19 & 13.14\pm10.64\\
\end{tabular*}
\label{tab:MinMaxJoints}
\end{table}

%%%%%%%%%%%%%%%%%%%%%%%%%%%%%%%%%%%%%%%%%%%%%%%%%%%%%%%%%%%%
%% Max Travelled Distance of Hololens in [cm]
%Max Movement in Level 0 - Avatar: 31.274 +- 6.748  
%Max Movement in Level 0 - no Avatar: NaN +- NaN  

%Max Movement in Level 1 - Avatar: 35.938 +- 10.596  
%Max Movement in Level 1 - no Avatar: 35.938 +- 9.251  

%Max Movement in Level 2 - Avatar: 38.255 +- 11.328  
%Max Movement in Level 2 - no Avatar: 34.914 +- 7.488  

%Max Movement in Level 3 - Avatar: 39.072 +- 12.081  
%Max Movement in Level 3 - no Avatar: 36.777 +- 7.945  

%Max Movement avg.- Avatar: 36.13 +- 10.45  
%Max Movement avg.- no Avatar: 35.88 +- 8.01  

%% Absolute travelled distance fo HoloLens2 in [m]
%CumSum Movement in Level 0 - Avatar: 4.508 +- 1.197  
%CumSum Movement in Level 0 - no Avatar: NaN +- NaN  

%CumSum Movement in Level 1 - Avatar: 1.975 +- 0.498  
%CumSum Movement in Level 1 - no Avatar: 2.269 +- 0.750  

%CumSum Movement in Level 2 - Avatar: 3.511 +- 1.034  
%CumSum Movement in Level 2 - no Avatar: 3.099 +- 0.735  

%CumSum Movement in Level 3 - Avatar: 5.266 +- 2.167  
%CumSum Movement in Level 3 - no Avatar: 4.025 +- 1.152  

%CumSum Movement avg.- Avatar: 3.13 +- 1.13  
%CumSum Movement avg.- no Avatar: 3.82 +- 1.81  
%%%%%%%%%%%%%%%%%%%%%%%%%%%%%%%%%%%%%%%%%%%%%%%%%%%%%%%%%%%%

\begin{table}
\tabcolsep=0.09cm
\caption{Total distance of head movement per level with (w) and without (w/o) a virtual player model (avatar). Given are the average values and standard deviation per run.}
\begin{tabular}{rcccc}
	 Avatar		    & 	Level 1             &		Level 2         & 	Level 3	            & Average\\ \midrule
    %\textbf{y}      & \SI{35.94(10.60)}{}   & \SI{38.26(11.36)}{}   & \SI{39.07(12.08)}{}   &  \textbf{\SI{36.13(10.45)}{}}\\
    %\textbf{ n} 	&  \SI{35.94(09.25)}{}  & \SI{34.91(07.49 )}{}  & \SI{36.78(07.95)}{}   & \textbf{\SI{35.88(08.01)}{}}\\
    \textbf{w}      & \SI{1.98(49)}{\meter}   & \SI{3.51(103)}{\meter}   & \SI{5.27(216)}{\meter}   &  \textbf{\SI{3.82(181)}{\meter}}\\
    \textbf{w/o} 	&  \SI{2.27(75)}{\meter}  & \SI{3.10(74)}{\meter}  & \SI{4.03(115)}{\meter}   & \textbf{\SI{3.13(113)}{\meter}}\\
\end{tabular}
\label{tab:HeadVsAvatar}
\end{table}

\section{Conclusion}
We present a VR musculoskeletal training application that can stimulate movement by a downhill skiing scenario. Our results show that upper body movement as well as individual joint angles can be correlated with data recorded by the HoloLens 2. Additionally, level parameters directly influence the intensity of the training. Therefore, we argue that the creation of individualized automatic training scenarios based on our downhill skiing scenario is feasible.

% \begin{acknowledgement}
%   ...
% \end{acknowledgement}
\vspace{-0.1cm}

\subsection*{Author Statement}
Research funding: This work was partially funded by the Interdisciplinary Competence Center for Interface Research (ICCIR) and the Forschungszentrum Medizintechnik Hamburg (FMTHH; grant 01fmthh2019). Conflict of interest: Authors state no conflict of interest. Informed consent: Informed consent has been obtained from all individuals included in this study. Ethical approval: The research related to human use complies with all the relevant national regulations, institutional policies and was performed in accordance with the tenets of the Helsinki Declaration, and has been approved by the authors' institutional review board or equivalent committee.

\vspace{-0.2cm}

\bibliographystyle{elsevier3}
\bibliography{ref}
\end{document}